\documentclass{article} 
\usepackage{iclr2023_conference,times}

\usepackage{hyperref}
\usepackage{url}           
\usepackage{booktabs}       
\usepackage{amsfonts}       
\usepackage{nicefrac}       
\usepackage{microtype}      
\usepackage{xcolor}         
\usepackage{amsmath,amssymb}
\usepackage{algorithmic}
\usepackage{graphicx}
\usepackage{textcomp}
\usepackage{cases}
\usepackage{overpic}
\usepackage{caption}
\usepackage{subcaption}
\usepackage{multicol, multirow}
\usepackage{float}
\usepackage{tabularx}
\usepackage{enumitem}
\usepackage{array}
\usepackage{ragged2e}
\usepackage{adjustbox}
\usepackage{wrapfig}
\usepackage{lipsum}

\def\by{{\bf y}}
\def\bz{{\bf z}}

\def\0{{\bf 0}}
\def\1{{\bf 1}}

\def\dist{\mbox{dist}}

\def\ie{i.e.}

\global\long\def\dist{\mathsf{d}}
\global\long\def\bdist{\hat{\mathsf{d}}}

\usepackage{flushend}

\title{Imbalanced Domain Generalization for Robust Single Cell Classification in Hematological Cytomorphology}

\author{%
 Rao Muhammad Umer$^1$,
 Armin Gruber$^{1,2}$,
 Sayedali Shetab Boushehri$^{1,3}$,
 Christian Metak$^1$, \\
 \textbf{Carsten Marr}$^1$\\
 \texttt{\{christian.metak, carsten.marr\}@helmholtz-muenchen.de}\\
 Institute of AI for Health, Helmholtz Zentrum München - German Research Center for \\ Environmental Health, Neuherberg 85764, Germany.$^{1}$ \\
 Laboratory of Leukemia Diagnostics, Department of Medicine III, University Hospital,\\ LMU Munich, Munich, Germany.$^{2}$ \\
 Data Science, Pharmaceutical Research and Early Development Informatics (pREDi),\\ Roche Innovation Center Munich, Germany.$^{3}$ 
}

\iclrfinalcopy 
\begin{document}

\maketitle

\begin{abstract}
   Accurate morphological classification of white blood cells (WBCs) is an important step in the diagnosis of leukemia, a disease in which nonfunctional blast cells accumulate in the bone marrow. Recently, deep convolutional neural networks (CNNs) have been successfully used to classify leukocytes by training them on single-cell images from a specific domain. Most CNN models assume that the distributions of the training and test data are similar, i.e., the data are independently and identically distributed. Therefore, they are not robust to different staining procedures, magnifications, resolutions, scanners, or imaging protocols, as well as variations in clinical centers or patient cohorts. In addition, domain-specific data imbalances affect the generalization performance of classifiers. Here, we train a robust CNN for WBC classification by addressing cross-domain data imbalance and domain shifts. To this end, we use two loss functions and demonstrate their effectiveness in out-of-distribution (OOD) generalization. Our approach achieves the best F1 macro score compared to other existing methods and is able to consider rare cell types. This is the first demonstration of imbalanced domain generalization in hematological cytomorphology and paves the way for robust single cell classification methods for the application in laboratories and clinics.
\end{abstract}

\section{Introduction}
Hematology deals with the study of blood, blood-forming tissue, and blood-related diseases. Precise and early diagnosis of a hematologic disorder is crucial for the successful treatment. For decades, microscopic examination and classification of blood cells in stained peripheral blood (PB) and bone marrow (BM) samples have been a key step for the diagnosis of hematological malignancies. Cytomorphologic examination using light microscopy~\citep{walter2022artificial} remains one of the backbones of hematological diagnostics, often representing the first step in the workup and guiding additional methods such as immunophenotyping, cytogenetics, and molecular genetics. During morphologic examination, blood samples are evaluated microscopically by hematologists and screened for the presence of atypical cells populations that can indicate conditions such as leukemia. Typically, at least 200 cells per sample have to be classified according to current clinical guidelines. Such manual evaluation and classification under the microscope can be tedious, repetitive, and time-consuming. It furthermore relies heavily on trained and experienced staff, and is prone to variabilities due to the human factor \ie, limited intra-/inter-observer reproducibility.

\begin{figure}[t!]
\centering
\includegraphics[width=1.0\linewidth]{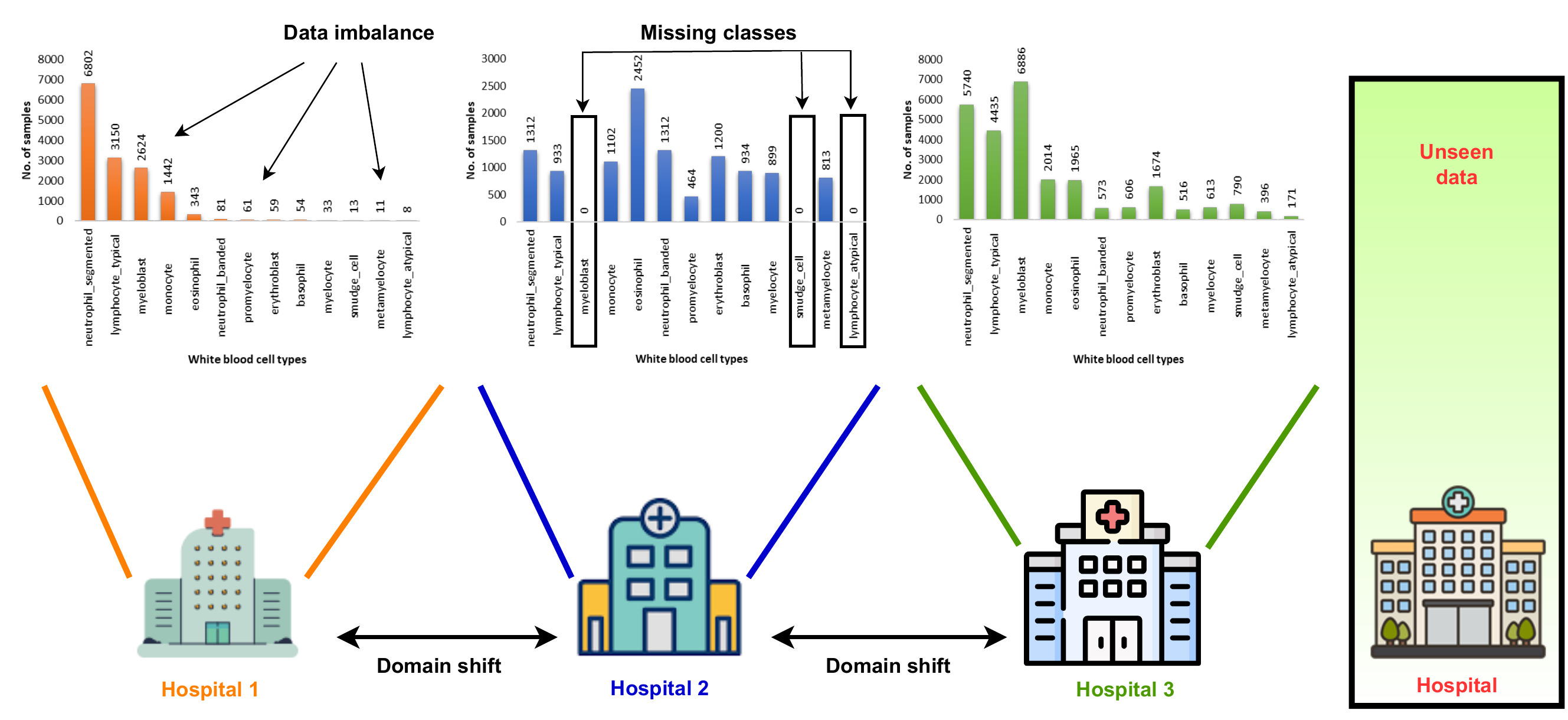}
\vspace{-0.3cm}
\caption{Key challenges for robust classification in an unseen target domain are data imbalance, missing classes, and domain shifts. In our application, hospital 1, 2, and 3 provide single white blood cell datasets from distinct domains with different class distributions.}
\label{fig:teaser}
\vspace{-0.3cm}
\end{figure}

Deep learning (DL) models have shown great potential in solving real-world classification tasks in various areas, such as computer vision~\citep{he2016resnet}, natural language processing~\citep{NEURIPS2020_1457c0d6}, and healthcare including medical imaging~\citep{matek2019cnn,wang2019deep} and histopathology~\citep{komura2018machine,hagele2020resolving}. They are usually developed and tested under the implicit assumption that the train and test data are drawn independently and identically from the same distribution (IID). Generalizing to unseen test domains is natural to humans, but challenging to machines. Domain generalization (DG) aims at training for domain invariant models robust against distribution shifts by utilizing data from distinct domains~\citep{zhou2022domaingensurvey,wang2022generalizing}. However, real-world data from multiple distinct domains often exhibit imbalanced label distributions \ie, a few classes contain a very large number of samples in the one domain, while only a few samples or none in another domain (see Fig.~\ref{fig:teaser}). Moreover, heavily imbalanced data distributions are common, and minority class samples in one domain could be abundant in other domains. Therefore, tackling the problem of cross-domain data imbalance is key to develop generalizable and diagnostically reliable models.  

Recently, several deep-learning models for the classification of WBCs have been proposed~\citep{sidhom2021deep,cheuque2022efficient,eckardt2022deep,salehi2022unsupervised,matthias2023expai}. In~\cite{matek2019cnn}, the authors trained a deep CNN model for the classification of single cell images from peripheral blood smears. The training data used for the model consisted of expert-annotated single cell images from the different subtypes of acute myeloid leukaemia (AML) patients. The images were processed using a ResNeXt-50 model~\citep{xie2017aggregated}, which provided high precision and recall for most diagnostically relevant classes, by assigning the class with the highest prediction probability to each image. So far, all existing state-of-the-art (SOTA) DL approaches are based on homogeneous datasets (identical train and test distribution), making the assumption of data balance during training by oversampling the minority classes or some loss re-weighting techniques~\citep{zhou2022domaingensurvey}. These datasets tend to be imbalanced not only due to varying cell distributions between different disease subtypes, but also due to varying patient populations and disease prevalences at different centers, making it difficult to predict the performance of a model on under-represented cases. This can result in limited performance and carries the risk of poor diagnostic performance. Additionally, single cell images from different labs can vary in sharpness, brightness, contrast, scale, color, and other properties, thus making it necessary to develop a robust classifier that can confidently classify single cell images regardless of their source domain and preanalytic handling.

\begin{figure}[t!]
\centering
\includegraphics[scale=1.0]{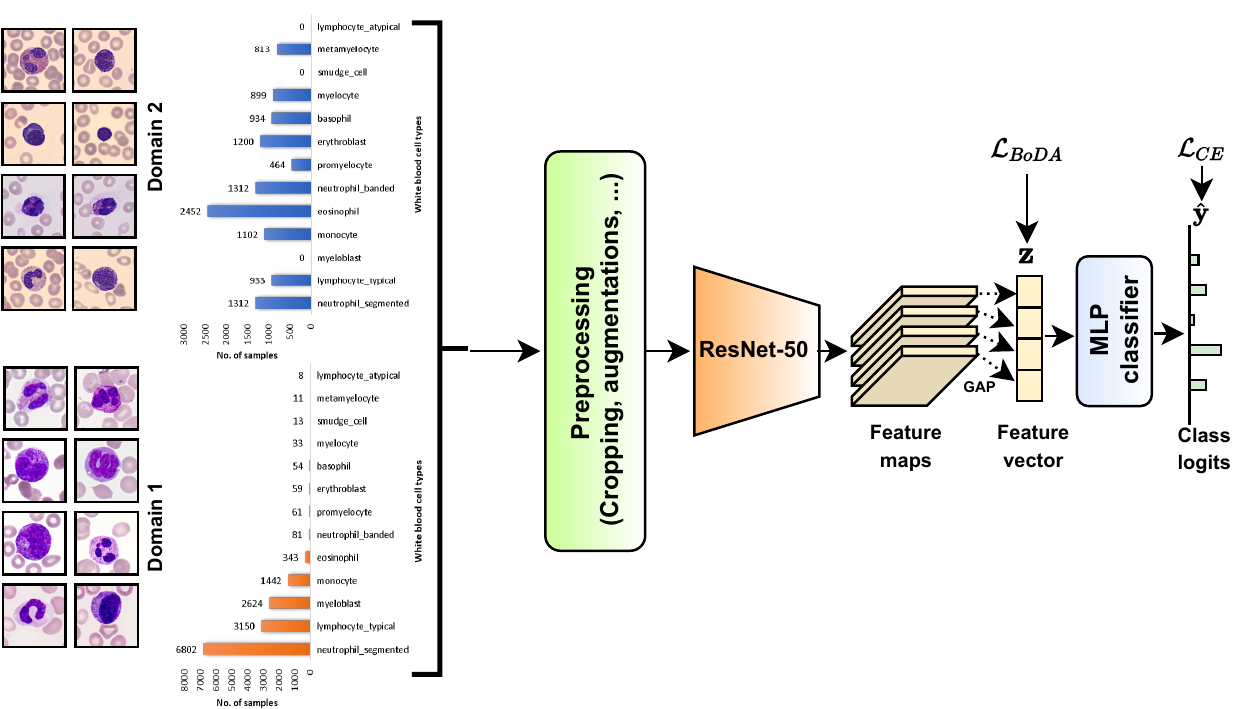}
\caption{Setup of our robust WBC classification approach: We take the single cell images from two different domains, preprocess with random resized cropping and various data augmentations, feed them into a pre-trained ResNet-50 network, and get the feature maps and the feature vector ($\bz$) by using global average pooling (GAP). The feature vectors are fed into the multilayer perceptron (MLP) classifier to get the class logits ($\hat{\by}$). The total loss consists of two terms ($\mathcal{L}_{\mathrm{BoDA}}$, $\mathcal{L}_{\mathrm{CE}}$) and is minimized during the fine-tuning of the network in an end-to-end manner. }
\label{fig:cls-setup}
\vspace{-0.3cm}
\end{figure}

Since single cell image data often originate from multiple distinct domains (\ie, hospitals and laboratories), many challenges arise (refer to section-\ref{sec:robust-cls-challenges} for more details) for robust classification, as shown in Fig.~\ref{fig:teaser}. The first challenge is the data imbalance within and across domains, the second one is the domain shift within and across domains, and the third one is missing class samples in certain domains. Our aim is to train a robust classifier, learning invariant features among multiple distinct domains, with each domain having its own domain shift and imbalanced label distribution problems, and generalizing to an unseen test target domain. To deal with the above problems, we train a robust classifier (as shown in Fig.~\ref{fig:cls-setup}) in an end-to-end fashion by minimizing the whole objective function~\eqref{eq:total_loss} inspired by the work of~\cite{yang2022boda}. The loss functions operate over the latent features and output layer respectively, and encourage similarity between features of the same class in different domains, and dissimilarity between features of different classes within and across domains, as well as correctly predicted class labels.  

\section{Methodology}
In this section, we explain robust classification challenges in detail and then propose a network training scheme to solve the existing problems for our task.

\subsection{Robust classification challenges}
\label{sec:robust-cls-challenges}
\subsubsection{Data imbalance}
Imbalanced data, where the number of samples in one class is significantly higher than the number in another class, can cause the learning algorithm to be biased towards the majority class. Data imbalance is an intrinsic problem~\citep{buda2018systematic,cao2019learning,yang2020rethinking} in real-world data, and it is even more severe in the medical domain. For example, the Matek\_19~\citep{matek2019cnn} dataset has severe class imbalance, where the majority class has more than 6500 samples, while the minority classes have less than 50 samples (Fig.~\ref{fig:teaser}). In the Acevedo\_20~\citep{acevedo2020dataset} dataset, we have somehow balanced data (Fig.~\ref{fig:teaser}), while in the INT\_20 dataset, we have a data imbalance problem with a long-tailed distribution (Fig.~\ref{fig:teaser}). 

\subsubsection{Domain shift}
In the medical domain, domain shifts can emerge due to different staining protocols, different scanners or acquisition protocols (\ie, background light, focus, etc.), different magnifications / resolutions, and variations in clinical centers or patient cohorts. The data can have distribution shifts even within a domain. For example, all three datasets used here have color variations due to different staining protocols and have different image resolutions due to different scanners (see Fig.~\ref{fig:cls-setup}). Furthermore, in the Acevedo\_20~\citep{acevedo2020dataset} dataset, as shown with visual image samples in Fig.~\ref{fig:cls-setup}, there are strong color variations within the domain.

\subsubsection{Missing classes}
In a domain~\ie, the Acevedo\_20~\citep{acevedo2020dataset} dataset (see Fig.~\ref{fig:teaser}), we have no single cell image data at all for certain classes. In some cases, divergent label distributions across domains can occur, rendering the problem more complex. Training data from different domain enhance classifier performance, where the minority samples or even no samples of the classes from one domain (hospital) are enriched with instances form the other domains.

\subsection{Network training losses}
For robust classifier learning, we train a deep CNN network (\ie,~ResNet-50~\citep{he2016resnet}) with standard hyperparameter settings as done in \cite{yang2022boda} by minimizing the following loss function:  
\begin{equation}
    \mathcal{L} = \underset{\boldsymbol{\theta}}{\arg \min }~\mathcal{L}_{\mathrm{CE}}+ \lambda\mathcal{L}_{\mathrm{BoDA}}.
    \label{eq:total_loss}
\end{equation}
Here, $\theta$ are the parameters of the network and $\lambda$ is a trade-off hyperparameter between the two loss terms.

For the $\mathcal{L}_{\mathrm{CE}}$ loss function, we use the standard cross-entropy loss applied to the output layer of the network ($\hat{\by}$) with the ground-truth label ($\by$). It is defined as:
\begin{equation}
    \mathcal{L}_{CE}(\hat{\by}, \by) = -\frac{1}{N} \sum_{n=1}^N \by_n \log \hat{\by}_n+\left(1 - \by_n\right) \log \left(1 - \hat{\by}_n\right)
    \label{eq:celoss}
\end{equation}
Here, $N$ is the mini-batch size and $\hat{\by}_n$ are the predicted class logits for image $n$.

For the $\mathcal{L}_{\mathrm{BoDA}}$ loss function, we use Balanced Domain-Class Distribution Alignment (BoDA)~\citep{yang2022boda} loss to tackle the data imbalance across domain-class $(d_i,c_i)$ pairs, which is applied to the latent features ($\bz$) as:
\begin{equation}
\label{eq:boda_loss}
\mathcal{L}_{BoDA}(\mathcal{\bz}, \boldsymbol{\psi})
= \sum\limits_{\mathbf{z}_i\in \mathcal{Z}} \frac{-1}{|\mathcal{D}|-1} \sum\limits_{d\in \mathcal{D}\setminus \{d_i\}} \log \frac{\exp{\left(- \boldsymbol{w}^{d,c_i}_{d_i,c_i} \bdist(\mathbf{z}_i, \boldsymbol{\psi}_{d,c_i})\right)}}{\sum_{(d',c') \in \mathcal{M} \setminus \{(d_i, c_i)\}} \exp{\left(- \boldsymbol{w}^{d',c'}_{d_i,c_i} \bdist(\mathbf{z}_i, \boldsymbol{\psi}_{d',c'})\right)}}.
\end{equation}
In Eq.~\eqref{eq:boda_loss}, the numerator represents positive cross-domain pairs distance $\bdist(.)$ that should be minimized (\ie, attract the same classes) during training, while the denominator represents negative cross-class pairs distance $\bdist(.)$ that should be maximized (\ie, separate different classes) during training. The $\boldsymbol{\psi}$ is mean of the feature vectors of domain-class pairs $(d, c)$. The $\boldsymbol{w}_{d, c}^{d^{\prime}, c^{\prime}}$ is the calibration parameter, which indicates how much to transfer $(d,c)$ to $(d^\prime, c^\prime)$ based on their relative sample size.
The distance $\dist$ can be set to the Euclidean distance $\dist(\mathbf{z}, \boldsymbol{\psi}_{d,c}) = \sqrt{(\mathbf{z} - \boldsymbol{\psi}_{d,c})^\top(\mathbf{z} - \boldsymbol{\psi}_{d,c})}$, which captures first-order statistics. To match higher-order statistics such as covariance, $\dist(\mathbf{z}, \{ \boldsymbol{\psi}_{d,c}, \boldsymbol{\Sigma}_{d,c} \}) = 1/N_{d_i,c_i} * \sqrt{(\mathbf{z} - \boldsymbol{\psi}_{d,c})^\top \boldsymbol{\Sigma}_{d,c}^{-1} (\mathbf{z} - \boldsymbol{\psi}_{d,c})}$ can be used, similar to the Mahalanobis distance~\citep{de2000mahalanobis}.
\begin{table}[t]
    \setlength{\tabcolsep}{10pt}
	\centering
	\caption{Imbalanced DG classification results (mean$\pm$std) determined by five-fold cross-validation on Acevedo\_20 and Matek\_19 validation-sets. Our base-line model is ResNet50, pretrained on ImageNet.}
	\vspace{-0.3cm}
	\resizebox{0.9\textwidth}{!}{
	\begin{tabular}{lcc}
        \toprule[1.5pt] 
       \textbf{Methods} & \textbf{F1-micro$\uparrow$} & \textbf{F1-macro$\uparrow$}  \\ 
        \midrule
		ERM~\citep{vapnik1999erm}   & \textbf{0.93} \scriptsize$\pm0.01$ & 0.77 \scriptsize$\pm0.02$ \\
		DANN~\citep{ganin2016dann}   & 0.87 \scriptsize$\pm0.03$ & 0.67 \scriptsize$\pm0.04$ \\
		CORAL (current SOTA DG)~\citep{sun2016coral}   & 0.92 \scriptsize$\pm0.01$ & 0.76 \scriptsize$\pm0.03$ \\
		Ours  & \textbf{0.93} \scriptsize$\pm0.01$ & \textbf{0.78} \scriptsize$\pm0.05$ \\
		Ours$^+$  & 0.90 \scriptsize$\pm0.02$ & 0.76 \scriptsize$\pm0.04$ \\
       \bottomrule[1.5pt]
    \end{tabular}}
	\label{tab:cls-res-comp-valset}
\end{table}

\begin{table}[t]
    \setlength{\tabcolsep}{10pt}
	\centering
	\caption{Imbalanced DG classification results (mean$\pm$std) determined by five-fold cross-validation on INT\_20 testset (unseen domain). Our base-line model is ResNet50, pretrained on ImageNet.}
	\vspace{-0.3cm}
	\resizebox{0.9\textwidth}{!}{
	\begin{tabular}{lcc}
        \toprule[1.5pt] 
       \textbf{Methods} & \textbf{F1-micro$\uparrow$} & \textbf{F1-macro$\uparrow$}  \\ 
        \midrule
		ERM~\citep{vapnik1999erm}   & 0.64 \scriptsize$\pm0.03$ & 0.40 \scriptsize$\pm0.05$ \\
		DANN~\citep{ganin2016dann}   & 0.59 \scriptsize$\pm0.07$ & 0.35 \scriptsize$\pm0.06$ \\
		CORAL (current SOTA DG)~\citep{sun2016coral}   & \textbf{0.66} \scriptsize$\pm0.03$ & 0.43 \scriptsize$\pm0.03$ \\
		Ours & \textbf{0.66} \scriptsize$\pm0.05$ & 0.43 \scriptsize$\pm0.06$ \\
		Ours$^+$  & 0.59 \scriptsize$\pm0.09$ & \textbf{0.46} \scriptsize$\pm0.08$ \\
       \bottomrule[1.5pt]
    \end{tabular}}
	\label{tab:cls-res-comp-testset}
	\vspace{-0.3cm}
\end{table}
\section{Experiments and Results} 
\subsection{Datasets and preprocessing}
For the robust classification task, we use three datasets of single cell peripheral blood images: Matek\_19~\citep{matek2019cnn}, Acevedo\_20~\citep{acevedo2020dataset}, and an internal data set (INT\_20). Appendix~\ref{appendix} provides detailed statistics of each dataset used in our experiments. We split the data into training ($80\%$ of the images) and validation ($20\%$ of the images) for the source domains (Acevedo\_20 and Matek\_19), and testset ($20\%$ of the images) for the unseen (OOD) test target domain (INT\_20). For a five-fold cross-validation, we perform random stratified splits of the single cell images into five folds. 

During data preprocessing, we crop all samples to the same size, so that in each dataset, the ratio of  cell pixel to background pixel is approximately identical. During training, we apply multiple standard transformations of the images, such as random resize, horizontal /  vertical flip, rotation, color-jitter, blur, and grayscale, to make the model more robust to image variations. 
\subsection{Evaluation Metrics and Results}
As the datasets are imbalanced, we aim to achieve a high F1-macro score on unseen test domain (OOD), which gives equal weight to each class regardless of its cardinality. Besides that, we also report the F1-micro score, which gives the weighted average of each class.
\begin{wrapfigure}[27]{r}{25em}
     \begin{subfigure}[b]{0.6\textwidth}
         \centering
         \includegraphics[width=0.75\linewidth]{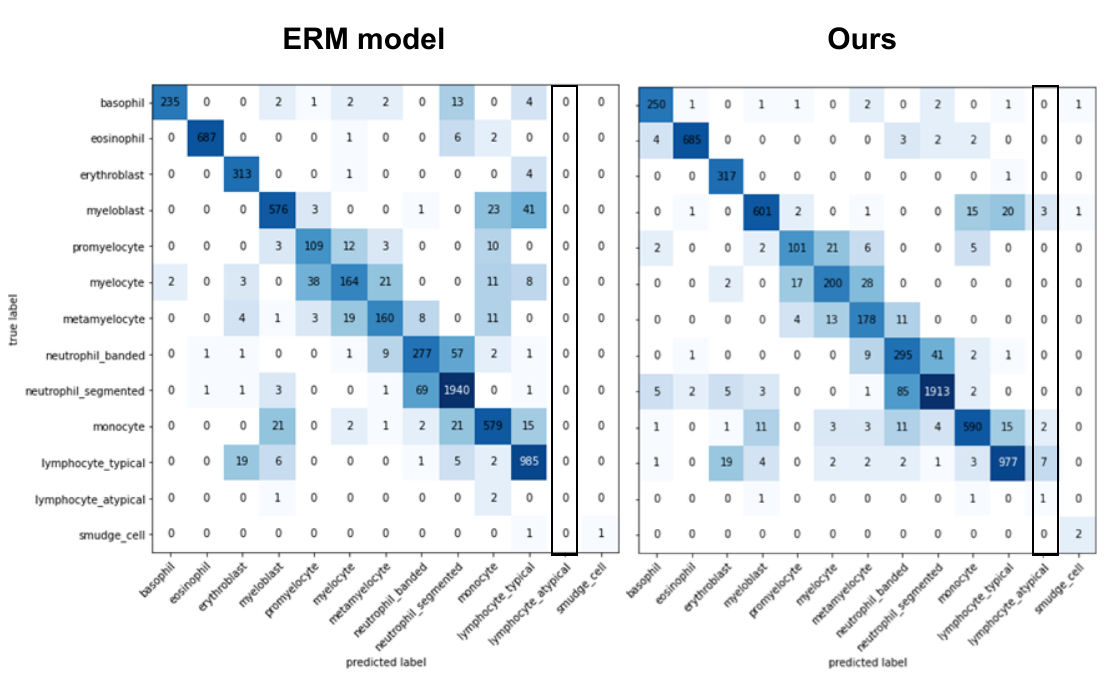} 
         \caption{Source (Matek\_19, Acevedo\_20) domain val-set}
         \label{fig:conf-mat-velset}
     \end{subfigure}
     \begin{subfigure}[b]{0.6\textwidth}
         \centering
         \includegraphics[width=0.75\linewidth]{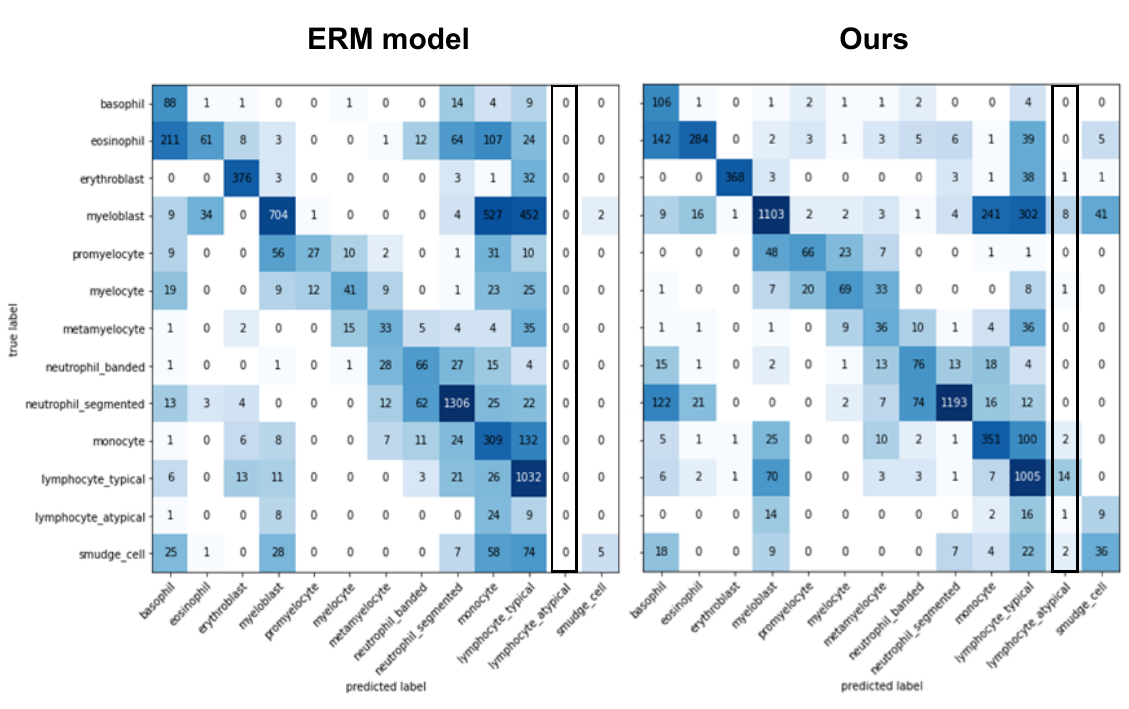}
         \caption{Target (INT\_20) domain testset (unseen)}
         \label{fig:}
     \end{subfigure}
    \caption{Confusion matrices show an improved classification of cells from the lowly populated \textbf{lymphocyte\_atypical} class with our method compared to the standard ERM model.}
    \label{fig:conf-mat-comp}
\end{wrapfigure}
We report imbalanced DG classification results in Table~\ref{tab:cls-res-comp-valset} and Table~\ref{tab:cls-res-comp-testset} determined by five-fold cross-validation on the source domains (Acevedo\_20 and Matek\_19) validation-set and unseen test domain (INT\_20). We compare our model to one vanilla training model~\eqref{eq:celoss} like ERM~\citep{vapnik1999erm}, and two current SOTA DG methods, such as DANN~\citep{ganin2016dann} and CORAL~\citep{sun2016coral}. We train two different variants of our model, one with coupled training of the encoder part (representation, $\bz$) and the MLP classifier part, and the other (denoted as Ours$^+$) with decoupled training of the encoder part and the classifier part by using class-balanced sampling. We achieve the best F1-marco score on the INT\_20 testset, while the DG method DANN performs worse than the vanilla ERM model due to its assumption of data balance during training. In Fig.~\ref{fig:conf-mat-comp}, we also compare the confusion matrix of our model with the ERM method. Our model performs better by recognizing samples from the minority class, while ERM misses samples from the minority class, such as lymphocyte\_atypical.  
\section{Conclusion}
We develop a robust CNN model for out-of-distribution generalization in hematological cytomorphology that tackles three main challenges: data imbalance, domain shifts, and missing classes. We show how existing pre-trained deep models can be improved for distinct domains by optimizing the loss function in the latent feature space and output logits of the network. Our work shows how biological, epidemiological, and technical variabilities in hematologic single WBC classification can be addressed for training robust AI-based cell classifiers, paving the way for their safe and productive use in a clinical setting.
\bibliographystyle{iclr2023_conference}
\bibliography{refs}

\newpage
\appendix
\section{Dataset Details}
\label{appendix}
In this section, we provide detailed information of the three datasets used in our experiments. Table~\ref{appendix:table:dataset-details} provides statistics and properties of Matek\_19, Acevedo\_20, and INT\_20.

The Matek\_19~\citep{matek2019cnn} dataset contains 14681 single cell images, divided into 13 classes, each image with a size of $400\times400\times3$ pixels corresponding to $29\times29$ micrometers. The resolution is $13.8$ pixels per micron.

The Acevedo\_20~\citep{acevedo2020dataset} dataset contains 11421 single cell images, divided into 10 classes, each image of $360\times363\times3$ pixels, corresponding to $36\times36.3$ micrometers. The resolution is $10$ pixels per micron.

INT\_20 is an internal dataset (currently not publicly available), which contains 26379 single cell images, divided into 13 classes, each image of size $288\times288\times3$ corresponding to $25\times25$ micrometers, so the resolution is $11.52$ pixels per micron.

\begin{table}[h!]
    \setlength{\tabcolsep}{5pt}
    \caption{Statistics and properties of the three datasets used in our experiments.}
    \vspace{-10pt}
    \label{appendix:table:dataset-details}
    \small
    \begin{center}
    \resizebox{0.8\textwidth}{!}{
    \begin{tabular}{lcccc}
    \toprule[1.5pt]
    \textbf{Dataset} & \textbf{\texttt{\#} classes} & \textbf{Image size} & \textbf{Image resolution} & \textbf{\texttt{\#} data samples} \\ \midrule
    \texttt{Matek\_19} & 13 & $400\times400\times3$ & {\begin{tabular}[c]{@{}c@{}}$29.0~\mu m\times29.0~\mu m $ \\ $=13.8$ pixels/micron\end{tabular}} & 14681 \\ \midrule
    \texttt{Acevedo\_20} & 10 & $360\times363\times3$ & {\begin{tabular}[c]{@{}c@{}}$36.0~\mu m\times36.3~\mu m $ \\ $=10$ pixels/micron\end{tabular}} & 11421  \\ \midrule
    \texttt{INT\_20} & 13 & $288\times288\times3$ & {\begin{tabular}[c]{@{}c@{}}$25.0~\mu m\times25.0~\mu m $ \\ $=11.52$ pixels/micron\end{tabular}} & 26379 \\
    \bottomrule[1.5pt]
    \end{tabular}}
    \end{center}
\end{table}

\begin{figure}
     \centering
     \begin{subfigure}[b]{1.0\textwidth}
         \centering
         \includegraphics[width=0.4\linewidth]{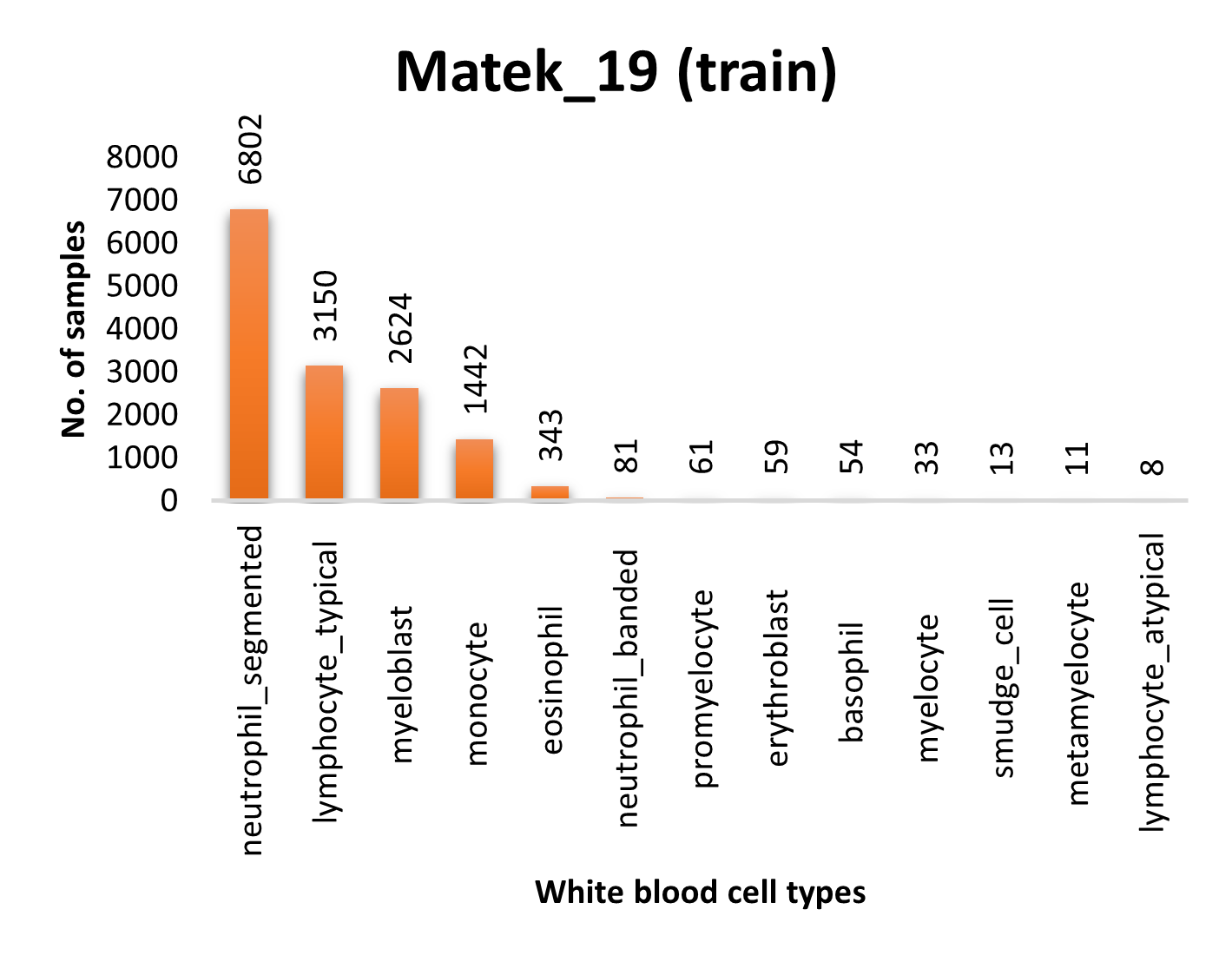} 
         \includegraphics[width=0.4\linewidth]{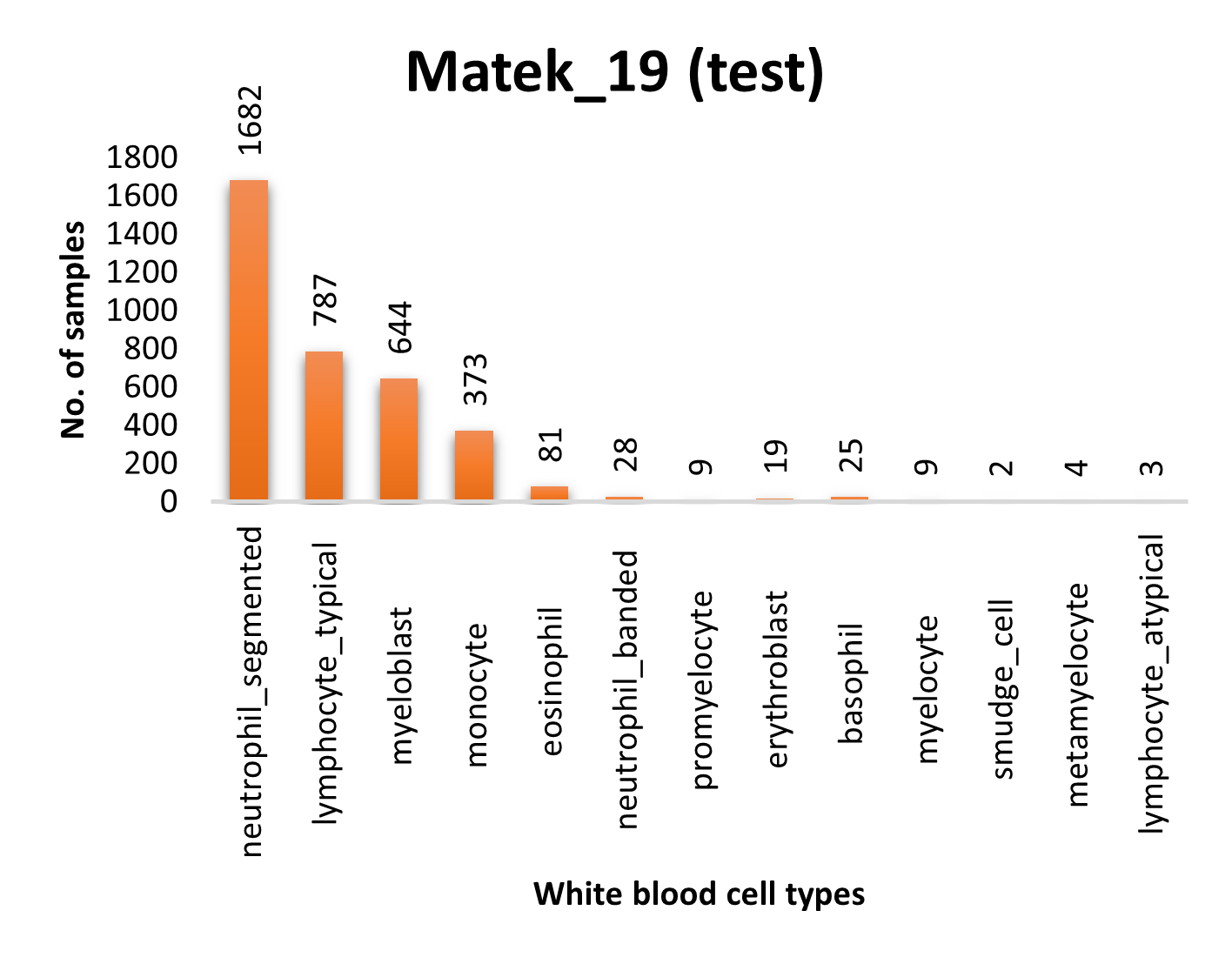}
         \caption{Matek\_19 train and test set}
         \label{fig:mat}
     \end{subfigure}
     \hfill
     \begin{subfigure}[b]{1.0\textwidth}
         \centering
         \includegraphics[width=0.4\linewidth]{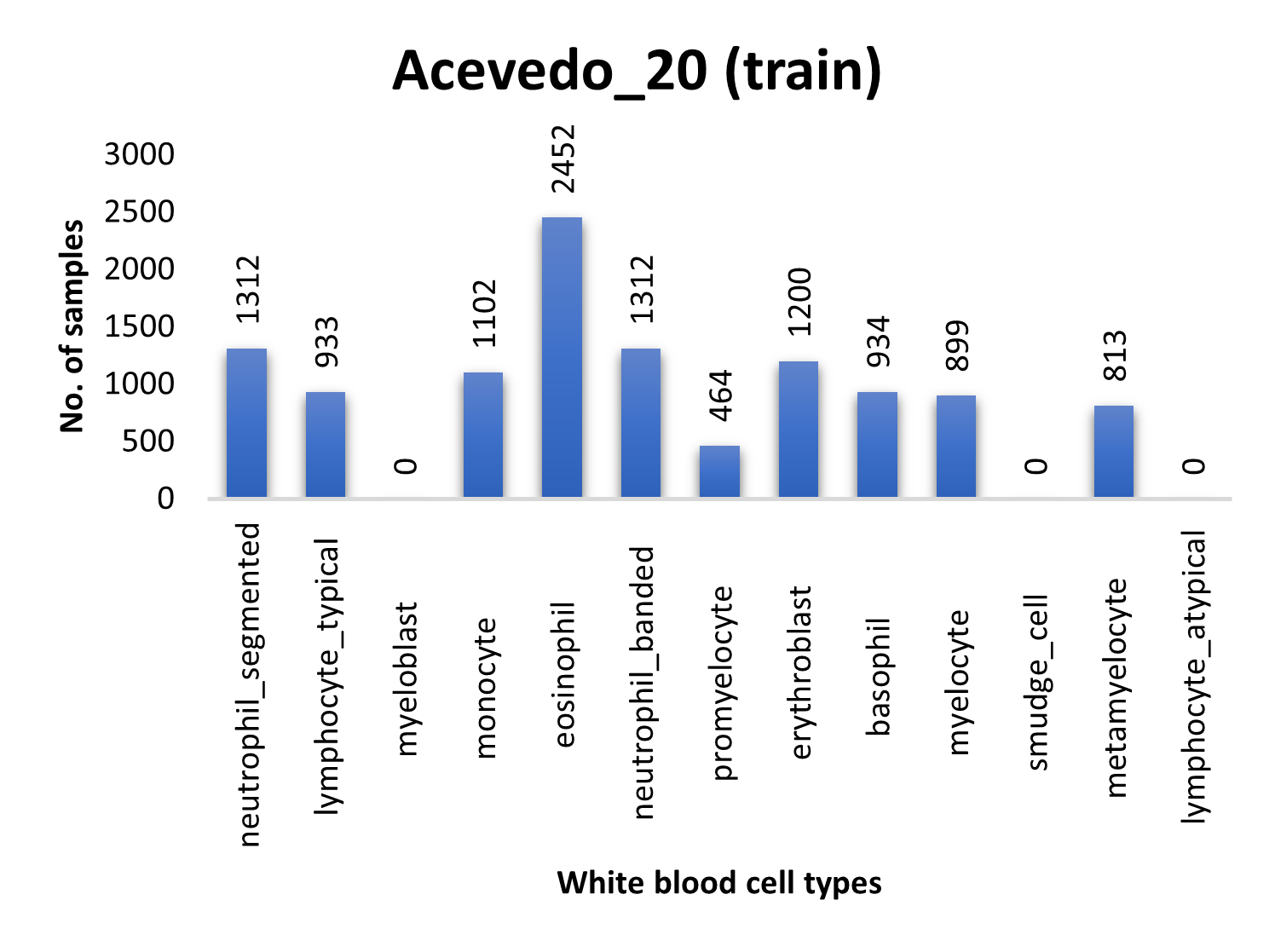}
         \includegraphics[width=0.4\linewidth]{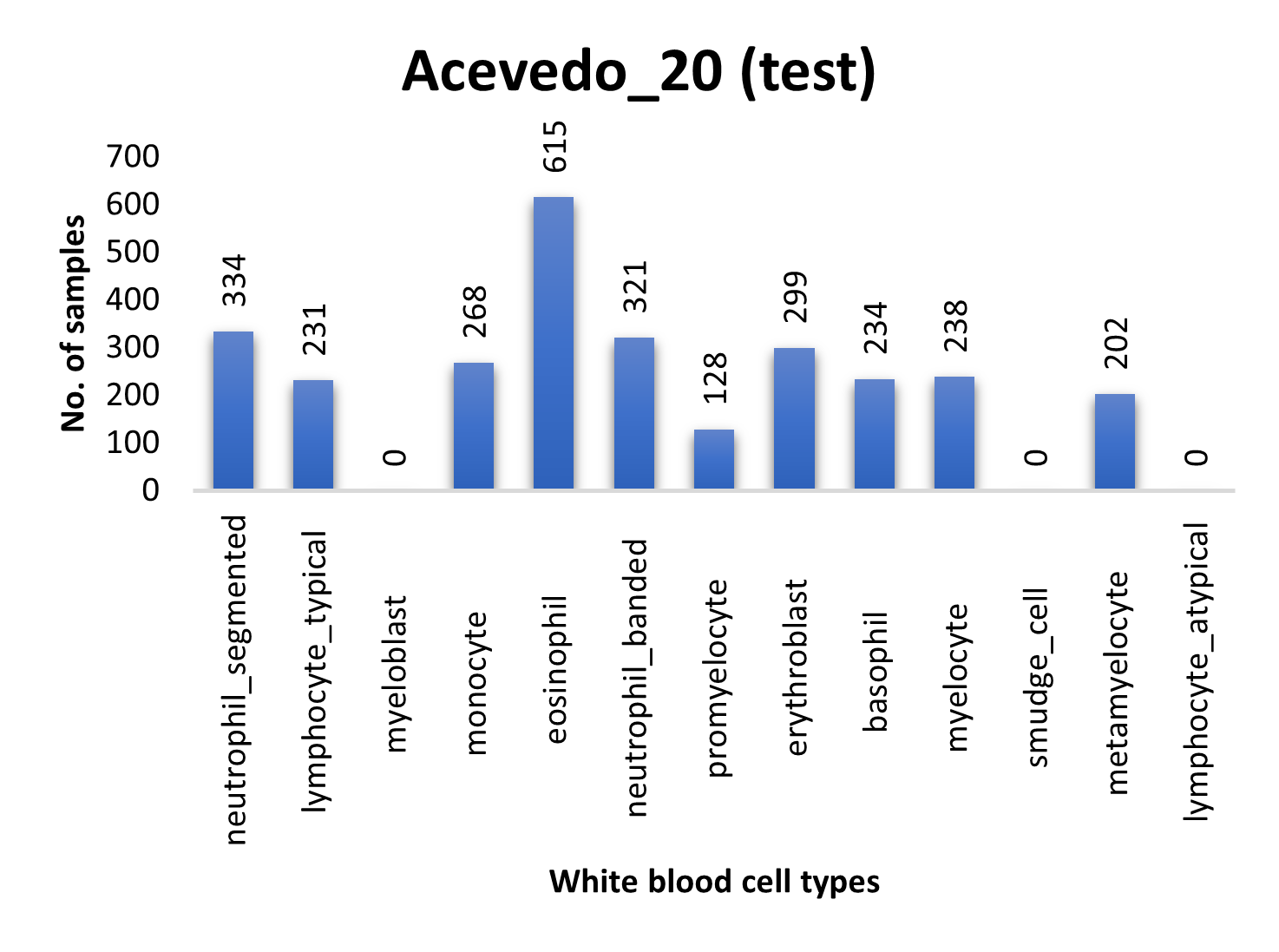}
         \caption{Acevedo\_20 train and test set}
         \label{fig:ace}
     \end{subfigure}
     \hfill
     \begin{subfigure}[b]{1.0\textwidth}
         \centering
         \includegraphics[width=0.4\linewidth]{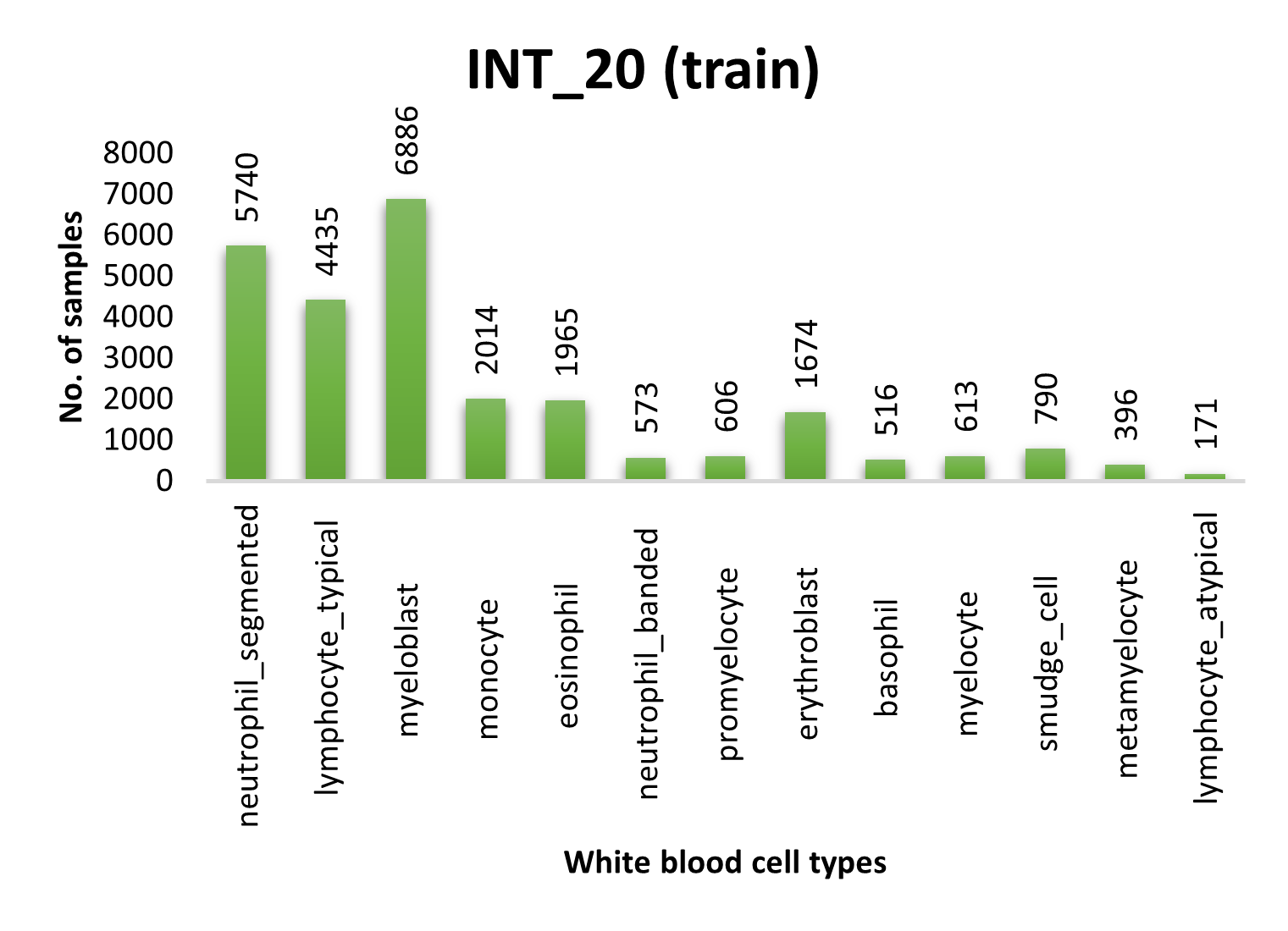}
         \includegraphics[width=0.4\linewidth]{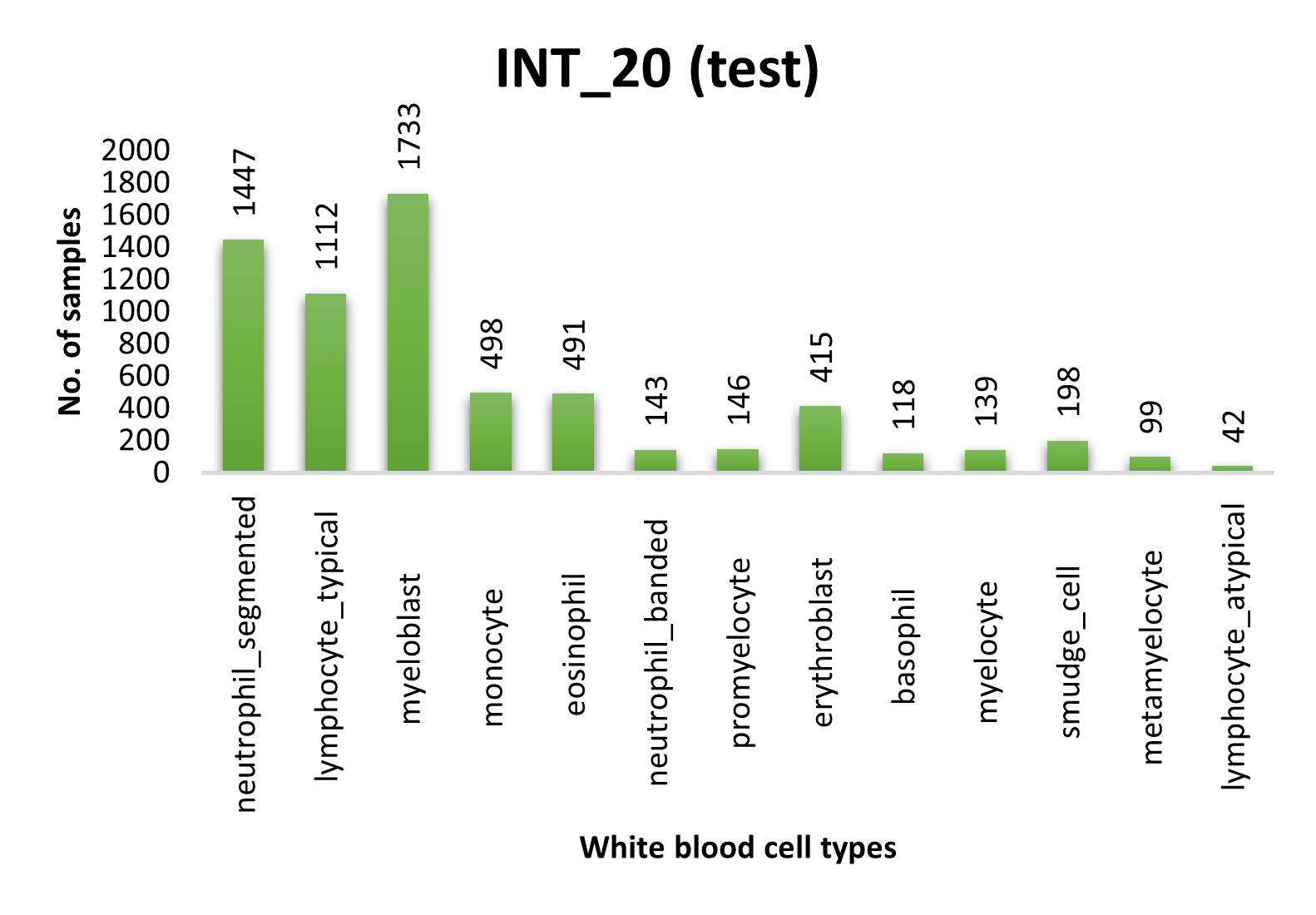}
         \caption{INT\_20 train and test set}
         \label{fig:mll}
     \end{subfigure}
        \caption{Train and test set class distributions of the three single cell datasets.}
        \label{appendix:fig:mat_ace_mll_dist}
\end{figure}

Fig.~\ref{appendix:fig:mat_ace_mll_dist} shows the sample distribution of each class in the trainset ($80\%$ of the single cell images) and the testset ($20\%$ of the single cell images) for all three datasets. We see highly imbalanced data in Matek\_19, missing samples in Acevedo\_20 domain, and a long-tail distribution in INT\_20.

\end{document}